\documentclass[letterpaper]{article} % DO NOT CHANGE THIS
\usepackage[preprint]{aaai2027}  % DO NOT CHANGE THIS
\usepackage[hyphens]{url}  % DO NOT CHANGE THIS
\usepackage{graphicx} % DO NOT CHANGE THIS
\usepackage{natbib}  % DO NOT CHANGE THIS AND DO NOT ADD ANY OPTIONS TO IT
\usepackage{caption} % DO NOT CHANGE THIS AND DO NOT ADD ANY OPTIONS TO IT
\usepackage{amsmath,amssymb}
\usepackage{multirow}
\usepackage{booktabs}
\usepackage{algorithm}
\usepackage{algorithmic}
\usepackage{tabularx}

\title{Beyond Relevance: Bayesian Evidence Acquisition for Agentic Whole-Slide Image Reasoning}
\author{
  Bryan Wong\textsuperscript{\rm 1,2}\thanks{This work is partially done during attachment with A*STAR.},
  Xun Xu\textsuperscript{\rm 2,3}\footnote{Corresponding authors.}\setcounter{aaai@corrmark}{\value{footnote}},
  Huazhu Fu\textsuperscript{\rm 2},
  Nancy F. Chen\textsuperscript{\rm 2,3},
  Mun Yong Yi\textsuperscript{\rm 1}\footnotemark[\value{aaai@corrmark}]
}
\affiliations{
  \textsuperscript{\rm 1}Korea Advanced Institute of Science and Technology (KAIST)\\
  \textsuperscript{\rm 2}Institute of Advanced Intelligence and Computing (IAIC), A*STAR \\
  \textsuperscript{\rm 3}Centre for Frontier AI Research (CFAR), A*STAR
}

\newcounter{aaai@corrmark}

\begin{document}

\maketitle

\begin{abstract}

Whole-slide image (WSI) reasoning requires an agent to sequentially acquire visual evidence before answering a diagnostic question. Existing training-free agentic frameworks formulate this process as iterative patch retrieval based on semantic relevance to the question. However, semantic relevance does not necessarily imply diagnostic informativeness in computational pathology, where competing diagnoses often exhibit similar and overlapping morphological patterns, making many patches semantically relevant yet diagnostically non-discriminative. Consequently, relevance-based retrieval may acquire redundant observations and leave diagnostic uncertainty unresolved. We propose \textbf{BEACON}, a plug-and-play agentic framework that reformulates WSI reasoning as a Bayesian evidence acquisition problem. BEACON maintains a probabilistic belief over competing diagnostic hypotheses and sequentially acquires patches by maximizing expected information gain (EIG) to reduce diagnostic uncertainty. An evidence controller then determines whether to answer, acquire additional evidence, or perform higher-resolution inspection. Built entirely from off-the-shelf foundation models, BEACON requires no additional training or fine-tuning. Extensive zero-shot experiments across five WSI-VQA benchmarks demonstrate that BEACON achieves the strongest overall performance among training-free agentic frameworks while substantially improving evidence acquisition efficiency, establishing Bayesian evidence acquisition as a principled paradigm for uncertainty-aware agentic WSI reasoning. The code is available at \url{https://github.com/bryanwong17/BEACON}.

\end{abstract}

\section{Introduction}
\label{sec:introduction}

Multimodal large language models (MLLMs) have recently demonstrated remarkable capabilities in vision-language reasoning by integrating visual perception with the reasoning ability of large language models (LLMs)~\cite{flamingo,llava}. These advances have substantially improved performance on multimodal reasoning tasks, including visual question answering (VQA)~\cite{vqa1,vqa2}. However, extending MLLMs to whole-slide images (WSIs) remains fundamentally challenging. Unlike conventional images, WSIs are gigapixel in scale and are typically divided into thousands of image patches for computational processing~\cite{patch_based1,patch_based2,clam}. Because current MLLMs cannot process all image patches simultaneously within their context window, WSI reasoning cannot rely on exhaustive slide observation.

\begin{figure}[t]
\centering
\includegraphics[width=\columnwidth]{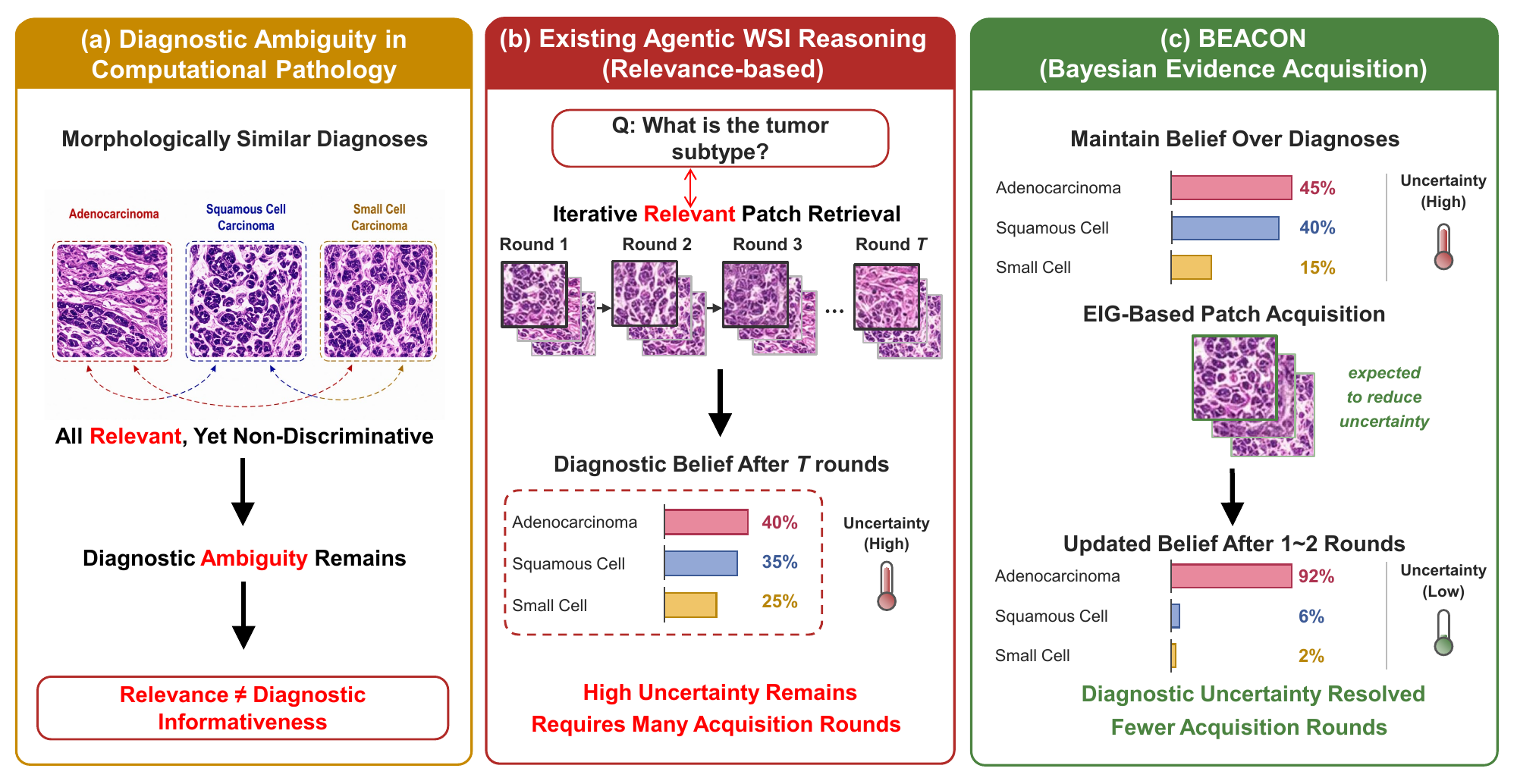}
\caption{(a) Morphologically similar diagnoses are all semantically relevant yet diagnostically non-discriminative, resulting in diagnostic ambiguity. (b) Existing agentic WSI reasoning frameworks iteratively retrieve relevant but non-discriminative patches, leaving diagnostic uncertainty unresolved. (c) BEACON acquires evidence by maximizing expected information gain (EIG), effectively reducing diagnostic uncertainty in fewer acquisition rounds.}
\label{fig:concept}
\end{figure}

Instead, WSI reasoning requires adaptively selecting which image patches to inspect in order to answer the input question, naturally formulating the problem as sequential decision making. Existing approaches can be broadly categorized into query-agnostic and query-aware methods. Query-agnostic methods rely on fixed, question-independent visual inputs, including downsampled thumbnail images, randomly sampled patches combined through majority voting, and aggregated patch features used by recent WSI-level MLLMs~\cite{slidechat,wsillava}. However, they cannot adaptively acquire the fine-grained morphological evidence required for diagnosis. In contrast, query-aware methods dynamically acquire visual evidence conditioned on the input question and intermediate reasoning state. More recently, training-free agentic frameworks, such as PathAgent~\cite{pathagent} and GIANT~\cite{giant}, iteratively retrieve image patches based on the input question and accumulated evidence, enabling adaptive exploration of gigapixel WSIs. Consequently, agentic frameworks have emerged as a promising paradigm for WSI reasoning.

Despite their different acquisition mechanisms, existing agentic frameworks largely rely on the same acquisition principle: selecting patches based on their semantic relevance to the question or previously acquired evidence. Although relevance-based strategies can efficiently reduce the search space, they implicitly assume that semantically relevant patches are diagnostically informative. This assumption often fails in computational pathology, where competing diagnostic hypotheses may exhibit similar and overlapping morphological patterns~\cite{high_overlap1,high_overlap2}, causing many patches to appear relevant while providing limited evidence for differentiating among them (Figure~\ref{fig:concept}(a)). Consequently, existing agentic frameworks may acquire relevant yet non-discriminative patches, leaving diagnostic uncertainty unresolved after multiple rounds of exploration (Figure~\ref{fig:concept}(b)).

We therefore argue that the central challenge in agentic WSI reasoning is not merely to narrow the search space based on semantic relevance but to identify observations that are expected to reduce diagnostic uncertainty. Motivated by this perspective, we formulate sequential patch acquisition as a Bayesian decision-theoretic problem~\cite{chaloner1995bayesian,rainforth2024modern}, maintaining a probabilistic belief over competing diagnostic hypotheses. Candidate patches are then evaluated according to their expected information gain (EIG)~\cite{lindley1956measure,houlsby2011bayesian}, thereby prioritizing observations that are most likely to resolve diagnostic uncertainty rather than those merely semantically relevant (Figure~\ref{fig:concept}(c)). 

Based on this formulation, we propose \textbf{BEACON} (\textbf{B}ayesian \textbf{E}vidence \textbf{A}cquisition for \textbf{CO}mputational pathology reaso\textbf{N}ing), a training-free, plug-and-play framework for agentic WSI reasoning that explicitly represents competing diagnostic hypotheses and performs uncertainty-aware patch acquisition. An evidence controller assesses the acquired evidence and determines whether it is sufficient to answer the question, whether additional patches should be acquired, or whether selected patches require higher-resolution inspection. Built entirely from off-the-shelf foundation models, BEACON requires no training or fine-tuning.

 Our key contributions are as follows:

\begin{itemize}

\item We formulate sequential patch acquisition in agentic WSI reasoning as a Bayesian evidence acquisition problem over competing diagnostic hypotheses and diagnostic uncertainty.

\item We propose \textbf{BEACON}, a training-free, plug-and-play agentic framework that performs uncertainty-aware patch acquisition by combining Bayesian belief estimation with expected information gain (EIG), without requiring additional training or fine-tuning.

\item Extensive zero-shot experiments across five WSI-VQA benchmarks and diverse foundation model backbones demonstrate that BEACON achieves the strongest overall performance among training-free agentic frameworks while requiring substantially fewer acquisition rounds.

\end{itemize}

\section{Related Work}
\label{sec:related_work}

\paragraph{Multimodal Models for WSI Analysis.}
Pathology VLMs such as PLIP~\cite{plip}, QuiltNet~\cite{quilt1m}, and CONCH~\cite{conch} provide patch-level embeddings for zero-shot retrieval, while pathology MLLMs, including Quilt-LLaVA~\cite{quiltllava}, PathGen-LLaVA~\cite{pathgen}, and Patho-R1~\cite{pathor1}, enable patch-level visual reasoning. At the slide level, ViLa-MIL~\cite{vilamil} and HiVE-MIL~\cite{hivemil} combine VLMs with MIL for slide-level prediction but are limited to classification and subtyping rather than general-purpose tasks such as VQA. Because entire WSIs cannot fit within the context window of current MLLMs, WSI-level methods rely on fixed, question-independent inputs, including thumbnails, randomly sampled patches, or aggregated patch embeddings projected into the LLM as in SlideChat~\cite{slidechat} and WSI-LLaVA~\cite{wsillava}. Consequently, question-specific evidence may never be observed, motivating adaptive patch acquisition at inference time.

\paragraph{Agentic Gigapixel WSI Reasoning.}
Agentic frameworks iteratively acquire visual evidence rather than operating on fixed inputs. Training-based methods, such as CPathAgent~\cite{cpathagent}, PathFinder~\cite{pathfinder}, and SlideSeek~\cite{slideseek}, learn pathology-specific navigation policies but require costly curated training data. In contrast, training-free methods, including PathAgent~\cite{pathagent} and GIANT~\cite{giant}, leverage off-the-shelf VLMs and MLLMs without additional training. PathAgent retrieves patches based on vision-language similarity, whereas GIANT predicts patch coordinates conditioned on previously observed evidence using an MLLM.

Despite these methodological differences, both approaches implicitly assume that semantically relevant observations are sufficient for effective evidence acquisition. This assumption is problematic in computational pathology, where competing diagnoses often exhibit subtle and overlapping morphological patterns. Consequently, semantically relevant patches may provide limited discriminative evidence and leave diagnostic uncertainty unresolved. Prior work in active learning~\cite{gal2017deep} and Bayesian optimal experimental design~\cite{chaloner1995bayesian,rainforth2024modern} has shown that uncertainty-aware acquisition can prioritize informative observations in sequential decision-making. However, these principles remain largely unexplored in gigapixel WSI reasoning. We therefore formulate patch acquisition as a Bayesian evidence acquisition problem targeting this expected reduction in diagnostic uncertainty.

\begin{figure*}[t]
\centering
\includegraphics[width=\textwidth]{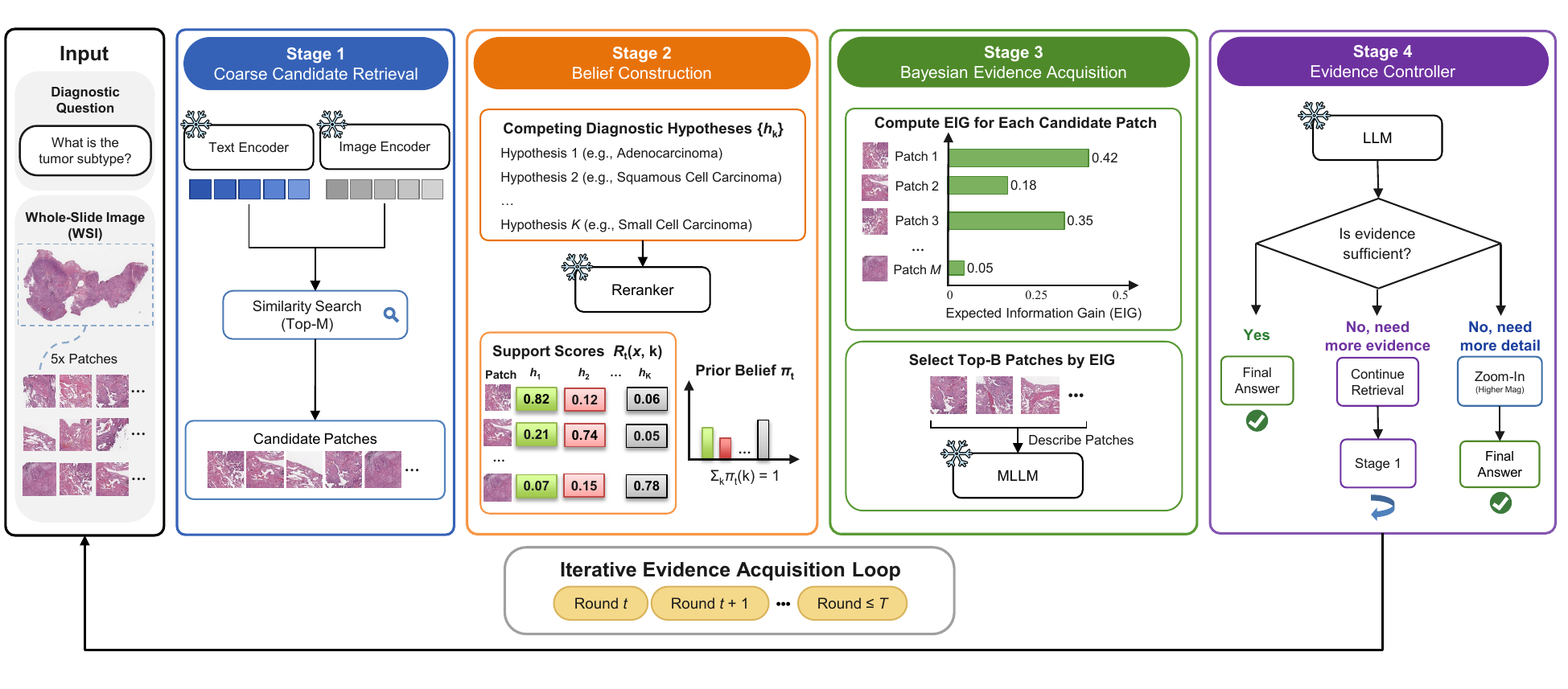}
\caption{Overview of the proposed BEACON agentic framework.}
\label{fig:framework}
\end{figure*}

\section{Methodology}
\label{sec:method}

\subsection{Overview of BEACON}
\label{sec:overview}

We formulate WSI reasoning as a Bayesian evidence acquisition problem. Unlike existing agentic frameworks that treat semantic relevance as the acquisition objective, BEACON uses relevance-based retrieval only to construct a compact candidate space and performs uncertainty-aware patch acquisition by maintaining a probabilistic belief over competing diagnostic hypotheses.

Each acquisition round consists of four stages. \textbf{(1) Coarse Candidate Retrieval} (Section~\ref{sec:retrieval}) uses a pathology VLM to restrict the search space to a compact candidate pool conditioned on the current retrieval query. \textbf{(2) Belief Construction} (Section~\ref{sec:belief}) estimates a probabilistic belief over competing diagnostic hypotheses from the retrieved candidates. \textbf{(3) Bayesian Evidence Acquisition} (Section~\ref{sec:eig}) evaluates candidate patches by their expected information gain and selects those providing the greatest expected uncertainty reduction. Finally, \textbf{(4) Evidence Control} (Section~\ref{sec:controller}) determines whether the accumulated evidence is sufficient to answer the question, whether higher-resolution inspection is required, or whether further exploration is needed. Figure~\ref{fig:framework} illustrates the overall framework, while the pseudocode is provided in the supplementary material.

\subsection{Problem Formulation}
\label{sec:problem}

A WSI is represented as a collection of image patches $\mathcal{X} = \{x_n\}_{n=1}^{N}$, where each patch $x_n$ is associated with spatial coordinates $c_n$. Given a diagnostic question $q$ and answer candidates $\mathcal{Y} = \{y_1, \ldots, y_K\}$, the objective is to predict the correct answer $y^\ast \in \mathcal{Y}$.

We treat WSI reasoning as sequential evidence acquisition. At round $t$, let $\mathcal{S}_t, \mathcal{U}_t$ denote the acquired and unobserved patch sets, with $\mathcal{S}_t \cap \mathcal{U}_t = \emptyset$, $\mathcal{S}_t \cup \mathcal{U}_t = \mathcal{X}$. Starting from $\mathcal{S}_1 = \emptyset$, BEACON iteratively acquires informative patches from $\mathcal{U}_t$, with each acquired patch yielding a question-specific description $\tilde{d}_n = f_{\mathrm{MLLM}}(x_n, q, \mathcal{Y}, c_n, \mathrm{mag})$, where $\mathrm{mag}$ is the patch magnification level.

\subsection{Stage 1: Coarse Candidate Retrieval}
\label{sec:retrieval}

Evaluating every unobserved patch under the Bayesian acquisition rule is computationally prohibitive for gigapixel WSIs. BEACON therefore first performs coarse candidate retrieval to efficiently reduce the search space by identifying patches that are semantically relevant to the current query before Bayesian evidence acquisition.

At round $t$, the retrieval query $r_t$ is initialized as the input question and iteratively refined by the evidence controller (Stage~\ref{sec:controller}). Let $\phi_T(\cdot)$ and $\phi_I(\cdot)$ denote the frozen text and image encoders of the VLM, respectively. Given the precomputed $5\times$ patch embeddings $e_x=\phi_I(x)\in\mathbb{R}^d$, the VLM encodes $r_t$ into a text embedding $\phi_T(r_t)\in\mathbb{R}^d$, and cosine similarity is computed as

\begin{equation} 
s_t(x) = \frac{\phi_T(r_t)^\top e_x}{\|\phi_T(r_t)\|_2 \, \|e_x\|_2}, \qquad x \in \mathcal{U}_t, 
\label{eq:retrieval}
\end{equation}

The top-$M$ patches are selected to form the candidate pool,
\begin{equation}
\mathcal{C}_t = \operatorname{TopM}_{x \in \mathcal{U}_t} \, s_t(x), \qquad M \ll |\mathcal{U}_t|, 
\label{eq:candidate}
\end{equation}

Subsequent Bayesian belief construction and evidence acquisition are performed exclusively over $\mathcal{C}_t$, substantially reducing computational cost while enabling uncertainty-aware patch selection.

\subsection{Stage 2: Belief Construction}
\label{sec:belief}

BEACON next constructs a probabilistic belief over the competing diagnostic hypotheses, providing the prior required for Bayesian evidence acquisition.

For each answer candidate $y_k \in \mathcal{Y}$, we define a diagnostic hypothesis $h_k$ asserting that $y_k$ is correct. Negation-style questions invert the hypothesis template, and composite choices (e.g., ``all of the above'') are expanded into logical combinations of their constituent hypotheses, yielding a mutually exclusive, collectively exhaustive set $\{h_1, \ldots, h_K\}$ over which a belief can be defined (see the supplementary material for details).

For every candidate patch $x \in \mathcal{C}_t$, a frozen reranker $f_{\mathrm{R}}$ estimates the support that the visual evidence provides for each hypothesis by contrasting its output logits for the affirmative (``yes'') and negative (``no'') response tokens:
\begin{equation}
    R_t(x,k) = \frac{\exp(\ell_{\mathrm{yes}}(d_x,h_k,q))}{\exp(\ell_{\mathrm{yes}}(d_x,h_k,q)) + \exp(\ell_{\mathrm{no}}(d_x,h_k,q))},
    \label{eq:reranker}
\end{equation}
where $d_x$ is the offline-generated description of patch $x$; $R_t(x,k)$ is close to $1$ when $x$ supports $h_k$ and close to $0$ when it contradicts it. Since these scores are not calibrated, we retain only their relative support via $L_1$ normalization,
\begin{equation}
    \bar{R}_t(x,k) = \frac{R_t(x,k)}{\sum_{\ell=1}^{K} R_t(x,\ell)}, \qquad \sum_{k=1}^{K} \bar{R}_t(x,k) = 1,
    \label{eq:norm}
\end{equation}
so that $\bar{R}_t(x,\cdot)$ is a categorical distribution over hypotheses for patch $x$. Aggregating over the candidate pool gives the belief,
\begin{equation}
    \pi_t(k) = \frac{\bar{\mu}_t(k)}{\sum_{\ell=1}^{K} \bar{\mu}_t(\ell)}, \quad s.t \quad \bar{\mu}_t(k) = \frac{1}{M} \sum_{x \in \mathcal{C}_t} \bar{R}_t(x,k),
    \label{eq:belief}
\end{equation}
where $M = |\mathcal{C}_t|$ denotes the number of candidate patches in the retrieved candidate pool, and $\pi_t(k)$ is the current belief that $h_k$ is correct, serving as the prior for the next stage.

\subsection{Stage 3: Bayesian Evidence Acquisition}
\label{sec:eig}

Given the prior belief $\pi_t$, BEACON selects the candidate patches expected to provide the greatest reduction in diagnostic uncertainty. Specifically, each candidate patch is evaluated according to the expected reduction in uncertainty that would result from inspecting it.

Let $Y \in \mathcal{Y}$ denote the unknown correct diagnosis, whose prior distribution is given by
\begin{equation}
    P(Y = y_k) = \pi_t(k).
    \label{eq:prior}
\end{equation}
For every candidate patch $x \in \mathcal{C}_t$, define a binary observation variable $O_x \in \{+,-\}$, where $O_x = +$ indicates that inspecting patch $x$ provides evidence supporting the correct diagnosis. The normalized hypothesis support is interpreted as the observation likelihood,
\begin{align}
    P(O_x = + \mid Y = y_k) &\triangleq \bar{R}_t(x,k), \nonumber \\
    P(O_x = - \mid Y = y_k) &= 1 - \bar{R}_t(x,k).
    \label{eq:likelihood}
\end{align}
Marginalizing over the prior yields the predictive probability of each observation,
\begin{align}
    P(O_x = +) &= \sum_{k=1}^{K} \bar{R}_t(x,k) \pi_t(k), \nonumber \\
    P(O_x = -) &= 1 - P(O_x = +).
    \label{eq:predictive}
\end{align}
Once patch $x$ is observed, the prior belief is updated according to Bayes' rule,
\begin{equation}
    P(Y = y_k \mid O_x) = \frac{P(O_x \mid Y = y_k)\, P(Y = y_k)}{P(O_x)},
    \label{eq:bayes}
\end{equation}
which yields the posterior beliefs
\begin{align}
    \pi_t(k \mid O_x = +) &= \frac{\bar{R}_t(x,k)\, \pi_t(k)}{P(O_x = +)}, \nonumber \\
    \pi_t(k \mid O_x = -) &= \frac{(1 - \bar{R}_t(x,k))\, \pi_t(k)}{P(O_x = -)}.
    \label{eq:posterior}
\end{align}

Diagnostic uncertainty \emph{before} observing $x$ is quantified by the entropy of the current belief, $H(\pi_t)$. Since the observation outcome is unknown in advance, we evaluate the expected posterior entropy \emph{after} observing $x$, weighted by the predictive probabilities in Eq.~\eqref{eq:predictive}. The expected information gain (EIG) is therefore
\begin{align}
    \mathrm{EIG}_t(x) ={} & H(\pi_t) - \Big[ P(O_x{=}{+})\, H\big(\pi_t(\cdot \mid O_x{=}{+})\big) \nonumber \\
    & + P(O_x{=}{-})\, H\big(\pi_t(\cdot \mid O_x{=}{-})\big) \Big],
    \label{eq:eig}
\end{align}
where $H(\pi_t) = -\sum_{k=1}^{K} \pi_t(k) \log \pi_t(k)$ denotes the Shannon entropy of the current belief. Equivalently,
\begin{equation}
    \mathrm{EIG}_t(x) = \underbrace{H(\pi_t)}_{\text{uncertainty now}} \;-\; \underbrace{\mathbb{E}_{O_x}\big[H(\pi_t(\cdot \mid O_x))\big]}_{\text{expected uncertainty after observing } x}.
    \label{eq:eig_expectation}
\end{equation}
By the chain rule of entropy, it can equivalently be written as
\begin{equation}
    \mathrm{EIG}_t(x) = H(Y) - H(Y \mid O_x) = I(Y; O_x),
    \label{eq:mi}
\end{equation}

where $I(Y; O_x)$ denotes the mutual information between the unknown answer hypothesis and the potential observation obtained from inspecting patch $x$, following the information-theoretic formulation of Bayesian experimental design~\cite{lindley1956measure,chaloner1995bayesian,houlsby2011bayesian}. This formulation shows that EIG quantifies the expected reduction in diagnostic uncertainty, so patches expected to provide greater discrimination among competing hypotheses receive higher acquisition scores.

At each round, the top-$B$ candidate patches with the highest EIG scores are selected for inspection, $\mathcal{B}_t = \operatorname*{TopB}_{x \in \mathcal{C}_t}\, \mathrm{EIG}_t(x)$. These patches are then passed to the MLLM to obtain question-specific descriptions, which constitute the newly acquired evidence for the current round, $\mathcal{E}_t=\left\{f_{\mathrm{MLLM}}(x,q,\mathcal{Y},c_x,\mathrm{mag}) \;\middle|\; x\in\mathcal{B}_t\right\}$, where $c_x$ and $\mathrm{mag}$ denote the spatial coordinates and magnification level of patch $x$, respectively.

\begin{table*}[t]
\centering
\small
\setlength{\tabcolsep}{3pt}
\begin{tabular}{c c cccccc}
\toprule
\multirow[c]{2}{*}{\textbf{Mode}} & \multirow[c]{2}{*}{\textbf{Method}} & \textbf{ExpertVQA} & \textbf{SlideBench} & \textbf{TCGA} & \textbf{GTEx} & \textbf{PANDA} & \multirow[c]{2}{*}{\textbf{Avg.}} \\
 & & ACC & ACC & BACC & BACC & BACC & \\
\midrule
\multirow[c]{6}{*}{Thumbnail}
 & Qwen2.5-VL-Instruct-7B & 35.16 & 42.64 & 5.24 & 14.57 & 15.82 & 22.69 \\
 & LLaVA-v1.5-7B [CVPR`24] & 25.78 & 32.99 & 4.27 & 13.68 & 16.06 & 18.56 \\
 & LLaVA-Med-1.5-7B [NeurIPS`23] & 36.72 & 24.87 & 3.33 & 5.00  & 16.06 & 17.20 \\
 & MedGemma-1.5-4B  & 39.84 & 33.50 & 3.95 & 22.00 & 17.31 & 23.32 \\
 & Quilt-LLaVA-1.5-7B [CVPR`24] & 34.38 & 31.98 & 5.91 & 11.40 & 17.55 & 20.24 \\
 & Patho-R1-7B [AAAI`26]  & 39.06 & 30.96 & 14.09 & 16.59 & 18.88 & 23.92 \\
\midrule
\multirow[c]{6}{*}{Patch Majority Vote}
 & Qwen2.5-VL-Instruct-7B & 34.38 & 42.64 & 7.53 & 13.00 & 15.67 & 22.64 \\
 & LLaVA-v1.5-7B [CVPR`24] & 28.12 & 34.01 & 3.33 & 19.13 & 16.67 & 20.25 \\
 & LLaVA-Med-1.5-7B [NeurIPS`23] & 39.06 & 24.87 & 3.33 & 5.71  & 11.64 & 16.92 \\
 & MedGemma-1.5-4B  & 35.94 & 31.98 & 3.75 & 19.66 & 17.31 & 21.73 \\
 & Quilt-LLaVA-1.5-7B [CVPR`24] & 34.38 & 32.99 & 4.55 & 26.15 & 16.67 & 22.95 \\
 & Patho-R1-7B [AAAI`26] & 32.81 & 32.49 & 8.83 & 28.44 & 15.67 & 23.65 \\
\midrule
\multirow[c]{2}{*}{WSI Features}
 & SlideChat [CVPR`25]  & 38.89 & 70.56$^{*}$ & 4.35 & 5.00 & 16.67 & 27.09 \\
 & WSI-LLaVA [ICCV`25] & 44.53 & 56.85$^{*}$ & 13.72 & 18.97 & 16.86 & 30.19 \\
\midrule
\multirow[c]{3}{*}{Agentic (Patho-R1-7B)}
 & GIANT$^\ddagger$ & \underline{43.75} & \underline{53.81} & 4.77 & 17.45 & \underline{19.57} & 27.87 \\
 & PathAgent [ECCV`26] & 42.19 & 51.27 & \underline{8.14} & \underline{46.10} & 16.41 & \underline{32.82} \\
 & \textbf{BEACON (ours)} & \textbf{46.09} & \textbf{55.84} & \textbf{15.27} & \textbf{48.90} & \textbf{22.34} & \textbf{37.69} \\
\midrule
\multirow[c]{4}{*}{Agentic (GPT-5)}
 & GIANT (5 iters.) & 46.88 & 51.27 & 11.36 & 45.39  & 16.41 & 34.26 \\
 & GIANT (20 iters.) & \underline{54.69} & \underline{53.81} & 15.27 & 45.67 & \textbf{19.54} & 37.80 \\
 & PathAgent [ECCV`26] & \textbf{57.03} & 50.76 & \underline{21.20} & \underline{54.49} & \underline{18.21} & 
 \underline{40.34} \\
 & \textbf{BEACON (ours)} & 53.12 & \textbf{58.88} & \textbf{22.89} & \textbf{65.27} & 17.52 & \textbf{43.54} \\
\bottomrule
\end{tabular}
\caption{Zero-shot comparison on five MultiPathQA benchmarks. The best and second-best results are highlighted in \textbf{bold} and \underline{underlined}. $^{*}$: both methods are trained on in-domain (TCGA) data, reported for reference only. GIANT$^\ddagger$: Patho-R1-7B lacks multi-turn tool-calling, so we report single-turn reasoning.}
\label{tab:main_results}
\end{table*}

\subsection{Stage 4: Evidence Control}
\label{sec:controller}

We employ a frozen reasoning LLM $\Psi$ to adaptively regulate the evidence acquisition process based on the accumulated evidence, determining whether the current evidence is sufficient for diagnosis or whether additional evidence acquisition is required~\cite{pathagent}.

\begin{equation}
    (\mathrm{state}_t, r_{t+1}) = \Psi(\mathcal{E}_t, q, \mathcal{Y}),
    \label{eq:controller}
\end{equation}
where $\mathrm{state}_t \in \{\textsc{Sufficient}, \textsc{Continue}, \textsc{Zoom}\}$ denotes the control action and $r_{t+1}$ is the refined retrieval query. Stage~\ref{sec:eig} determines \emph{which} evidence to acquire, whereas the controller determines \emph{whether} additional acquisition is needed and the next action. The controller performs one of three actions:

\begin{itemize}
    \item \textbf{\textsc{Sufficient}:} The accumulated evidence is considered sufficient to answer the question, and the final answer is generated.

    \item \textbf{\textsc{Continue}:} The missing diagnostic evidence is summarized into a refined retrieval query $r_{t+1}$ to guide the next round of coarse candidate retrieval (Stage~\ref{sec:retrieval}).

    \item \textbf{\textsc{Zoom}:} Higher-resolution inspection is performed by subdividing an acquired patch, using the VLM to select an informative sub-patch, and obtaining an additional MLLM description to answer the question.
\end{itemize}

\section{Experiments}
\label{sec:experiments}

\subsection{Experimental Setup}
Unless otherwise stated, PLIP~\cite{plip} is used for coarse candidate retrieval, Qwen3-Reranker-4B~\cite{qwen3reranker} for belief construction, and Quilt-LLaVA~\cite{quiltllava} for offline generic patch description generation. For question-conditioned patch description and evidence control, we consider two backbone configurations in Table~\ref{tab:main_results}: (1) Patho-R1-7B~\cite{pathor1} with Qwen3-4B~\cite{qwen3}, which serves as the default setting, and (2) GPT-5~\cite{gpt5}, which replaces both components. Agentic baselines use the corresponding foundation model configurations for fair comparison. We fix the coarse candidate pool size to $M=25$, acquire $B=5$ patches per round, and set the maximum acquisition budget to $T=5$ across all experiments. We follow PathAgent~\cite{pathagent} for WSI preprocessing and patch extraction. All experiments are conducted on Ubuntu 20.04 using two NVIDIA RTX A5000 GPUs. GPT-5 API costs are estimated using OpenRouter pricing (\$1.25/\$10 per 1M input/output tokens) without prompt-caching discounts.

\subsection{Comparison with State-of-the-Art Methods}
\noindent\textbf{Datasets and Evaluation.} We evaluate BEACON on MultiPathQA~\cite{giant}, which comprises five WSI-VQA benchmarks with 932 WSI-question pairs covering five clinical tasks: TCGA ExpertVQA (128, expert-authored questions), TCGA SlideBench (197, adopted from SlideChat~\cite{slidechat}), TCGA~\cite{tcga} (221, 30-way cancer-type diagnosis), GTEx~\cite{gtex} (190, 20-way organ classification), and PANDA~\cite{panda} (196, 6-way ISUP grading). We report accuracy (ACC) for TCGA ExpertVQA and TCGA SlideBench, and balanced accuracy (BACC) for TCGA, GTEx, and PANDA. Prior work~\cite{pathagent} uses string-similarity matching to map model outputs to answer choices. To avoid format-dependent mismatches and ensure consistent evaluation, we use an LLM-based evaluator to infer the intended answer choice for all methods (see supplementary materials).

\noindent\textbf{Baseline Methods.} We compare BEACON against 10 baselines spanning three categories: (1) query-agnostic methods across six MLLM backbones~\cite{qwen2.5vl,llava1.5,llavamed,quiltllava,pathor1,medgemma1.5} under the Thumbnail and Patch Majority Vote settings, following prior work~\cite{slideseek,slidechat}; (2) WSI-level trained MLLMs (SlideChat~\cite{slidechat} and WSI-LLaVA~\cite{wsillava}); and (3) training-free agentic frameworks (PathAgent~\cite{pathagent} and GIANT~\cite{giant}).

\noindent\textbf{Main Results.} As shown in Table~\ref{tab:main_results}, BEACON achieves the strongest overall zero-shot performance among training-free agentic frameworks on MultiPathQA. Under the default Patho-R1-7B setting, BEACON achieves the best performance across all five benchmarks, outperforming PathAgent by 4.87\% on average. These gains transfer to GPT-5, where BEACON achieves the highest average performance, outperforming PathAgent by 3.20\% on average. Together, these results demonstrate the effectiveness of Bayesian evidence acquisition across both pathology-specific and general-purpose foundation models.

\subsection{Bayesian Evidence Acquisition}

\noindent\textbf{Acquisition Strategy.}
Table~\ref{tab:acquisition_ablation} compares Bayesian evidence acquisition against random sampling and relevance-based retrieval at the default acquisition budget ($B=5$). Bayesian evidence acquisition consistently outperforms both strategies across all benchmarks and pathology VLM backbones (PLIP~\cite{plip} and CONCH~\cite{conch}), with 2.03\%--6.25\% gains over relevance-based retrieval. These results suggest that semantic relevance alone is insufficient for selecting diagnostically informative evidence. We further investigate why semantic relevance is a poor proxy for diagnostic informativeness in Section~\ref{sec:why}.

\noindent\textbf{Component Ablation.}
Table~\ref{tab:component_ablation} removes each component in turn: (1) coarse retrieval (Stage~1), (2) EIG-based acquisition (Stage~3), (3) both belief construction and EIG (Stage~2+3), and (4) evidence control (Stage~4); see the supplementary material for detailed ablation settings. Removing any component degrades performance, indicating that all four stages contribute to BEACON's overall effectiveness. The largest drop occurs when both belief construction and EIG are removed, confirming that Bayesian belief modeling and uncertainty-aware evidence acquisition are the primary drivers of BEACON's performance.

\begin{figure}[t]
\centering
\includegraphics[width=0.75\columnwidth]{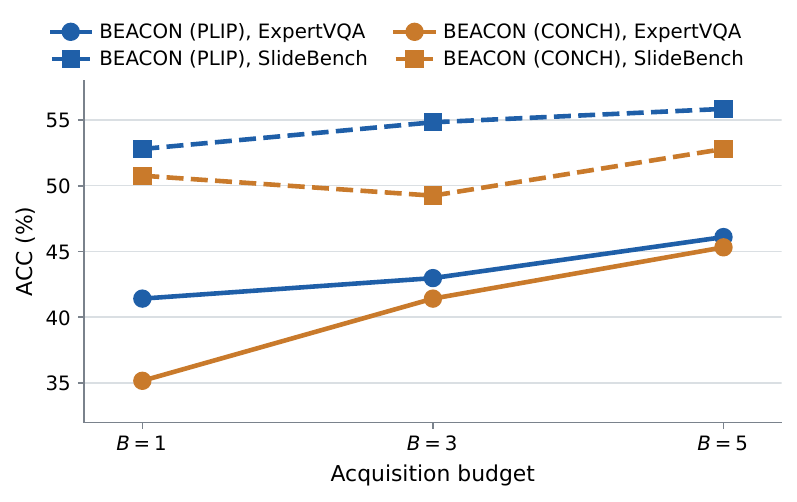}
\caption{Accuracy of the proposed Bayesian evidence acquisition rule as the per-round acquisition budget $B$ grows.}
\label{fig:acquisition_budget}
\end{figure}

\noindent\textbf{Effect of Acquisition Budget.}
Figure~\ref{fig:acquisition_budget} sweeps the per-round acquisition budget $B \in \{1,3,5\}$ across two VLMs and datasets. Larger acquisition budgets generally lead to higher accuracy, with the best performance achieved at $B=5$ across all settings, suggesting that EIG-based acquisition benefits from inspecting additional informative evidence.

\begin{table}[t]
\centering
\small
\setlength{\tabcolsep}{4pt}
\begin{tabular}{lccc}
\toprule
Method & ExpertVQA & SlideBench & PANDA \\
 & ACC & ACC & BACC \\
\midrule
Random & 38.28 & 49.23 & 15.67 \\
\midrule
PLIP & 39.84 & 51.78 & 19.98 \\
\textbf{BEACON (PLIP)} & \textbf{46.09} & \textbf{55.84} & \textbf{22.34} \\
\midrule
CONCH & 39.84 & 50.76 & 17.11 \\
\textbf{BEACON (CONCH)} & \textbf{45.31} & \textbf{52.79} & \textbf{23.40} \\
\bottomrule
\end{tabular}
\caption{Comparison of Bayesian evidence acquisition with random sampling and relevance-based retrieval under the default acquisition budget ($B=5$).}
\label{tab:acquisition_ablation}
\end{table}

\begin{table}[t]
\centering
\small
\setlength{\tabcolsep}{4pt}
\begin{tabular}{lcc}
\toprule
Method & ExpertVQA & SlideBench \\
 & ACC & ACC \\
\midrule
\textbf{BEACON (Full)} & \textbf{46.09} & \textbf{55.84} \\
\midrule
w/o Stage 1 (Coarse Retrieval) & 44.53 & 51.78 \\
w/o Stage 3 (EIG) & 45.31 & 53.30 \\
w/o Stage 2+3 (Belief \& EIG) & 39.84 & 51.78 \\
w/o Stage 4 (Evidence Control) & 44.53 & 52.28 \\
\bottomrule
\end{tabular}
\caption{Component ablation of BEACON by removing individual components.}
\label{tab:component_ablation}
\end{table}

\begin{figure*}[t]
\centering
\includegraphics[width=\textwidth]{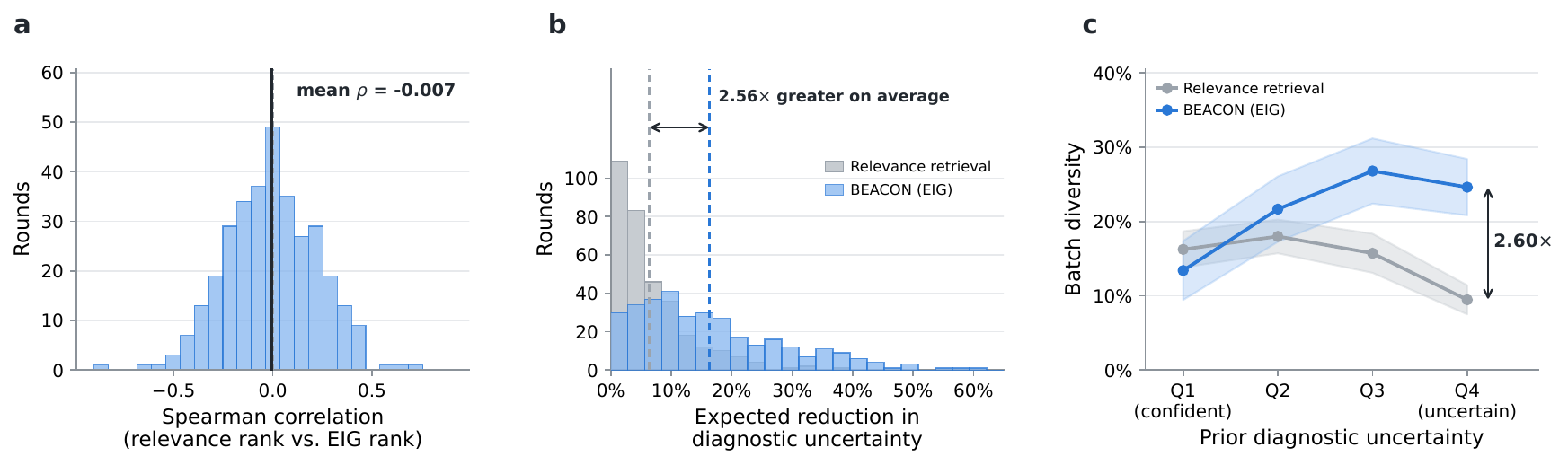}
\caption{Analysis of BEACON's acquisition strategy. (a) Relevance and EIG rankings are nearly independent. (b) EIG-selected patches achieve greater uncertainty reduction. (c) EIG provides larger gains as diagnostic uncertainty increases.}
\label{fig:BEACON_analysis}
\end{figure*}

\subsection{Why Does BEACON Work?}
\label{sec:why}

Table~\ref{tab:acquisition_ablation} shows that Bayesian evidence acquisition outperforms relevance-based retrieval. To understand why, we compare both strategies under identical backbones, candidate pools, and belief states, differing only in the patch selection criterion. Figure~\ref{fig:BEACON_analysis} summarizes the analysis over 329 acquisition rounds from ExpertVQA and SlideBench; detailed settings are provided in the supplementary material.

\noindent\textbf{Relevance Does Not Predict Diagnostic Informativeness.}
We rank the same candidate patches using cosine similarity (Eq.~\eqref{eq:retrieval}) and EIG (Eq.~\eqref{eq:eig}), which serves as our measure of diagnostic informativeness. Figure~\ref{fig:BEACON_analysis}(a) shows that the two rankings are essentially uncorrelated (mean Spearman $\rho=-0.007$ across 329 rounds), indicating that semantic relevance is a poor proxy for diagnostic informativeness.

\noindent\textbf{EIG Selects More Informative Evidence.}
We next ask whether explicitly optimizing for information gain leads to more informative evidence acquisition. Figure~\ref{fig:BEACON_analysis}(b) shows that EIG-selected patches achieve a $2.56\times$ larger expected uncertainty reduction than relevance-selected patches.

\noindent\textbf{The Advantage is Adaptive to Diagnostic Uncertainty.}
While EIG consistently selects more informative evidence, its benefit depends on the underlying diagnostic uncertainty. Stratifying acquisition rounds by prior diagnostic uncertainty $H(\pi_t)$, Figure~\ref{fig:BEACON_analysis}(c) shows that EIG's advantage becomes more pronounced as diagnostic uncertainty increases. In low-uncertainty cases, the two strategies achieve comparable evidential diversity (Q1: $13\%$ vs.\ $16\%$). However, when multiple diagnostic hypotheses remain plausible, EIG increasingly selects patches that better discriminate among them, achieving $2.60\times$ higher evidential diversity in the most uncertain cases (Q4: $25\%$ vs.\ $9\%$). This matches the Bayesian formulation: the benefit of EIG is naturally limited when the belief is concentrated, but increases as diagnostic uncertainty grows.

\subsection{Evidence Acquisition Efficiency}
\label{sec:efficiency}

\begin{figure}[tb]
\centering
\includegraphics[width=\columnwidth]{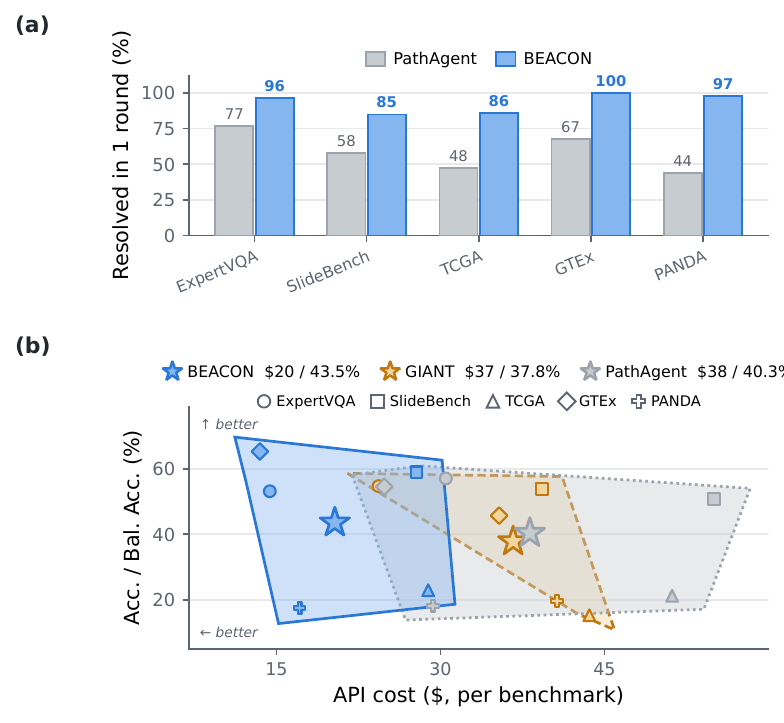}
\caption{(GPT-5 setting) (a) BEACON resolves more cases within a single acquisition round than PathAgent. (b) BEACON achieves the best cost--accuracy trade-off, outperforming PathAgent and GIANT ($\bigstar$: five-benchmark average cost/accuracy).}
\label{fig:efficiency_cost}
\end{figure}

Beyond accuracy, we evaluate whether informative evidence acquisition helps agents reach predictions more quickly.

\noindent\textbf{Acquisition Efficiency.}
BEACON resolves $85$--$100\%$ of cases within a single acquisition round across all benchmarks (Figure~\ref{fig:efficiency_cost}a), compared with only $44$--$77\%$ for PathAgent, indicating that EIG-based acquisition reaches confident decisions in fewer rounds.

\noindent\textbf{Cost--Accuracy Trade-off.}
By requiring fewer acquisition rounds, BEACON also achieves the best cost--accuracy trade-off across the five benchmarks (Figure~\ref{fig:efficiency_cost}b), reducing API cost by $46.6\%$ and $44.3\%$ compared with PathAgent and GIANT, respectively, while achieving higher average accuracy across all benchmarks (Table~\ref{tab:main_results}).

\subsection{Plug-and-play Compatibility}
Table~\ref{tab:backbone_ablation} evaluates BEACON across four backbone combinations of pathology VLMs (PLIP, CONCH) and MLLMs (Patho-R1-7B, PathGen-LLaVA). BEACON matches or outperforms PathAgent in 17 of 20 combinations, indicating that its gains arise from the acquisition strategy itself rather than any particular backbone pairing.

\begin{table}[t]
\centering
\small
\setlength{\tabcolsep}{2pt}
\begin{tabular}{llccccc}
\toprule
\textbf{VLM/MLLM} & \textbf{Method} & Exp. & Sli. & TCGA & GTEx & PANDA \\
\midrule
\multirow{2}{*}{P/R1}
& PathAgent & 42.19 & 51.27 & 8.14 & 46.10 & 16.41 \\
& \textbf{BEACON} 
& \textbf{46.09} 
& \textbf{55.84} 
& \textbf{15.27} 
& \textbf{48.90} 
& \textbf{22.34} \\
\midrule
\multirow{2}{*}{C/R1}
& PathAgent & 41.41 & 48.22 & 12.62 & 46.73 & \textbf{23.40} \\
& \textbf{BEACON} 
& \textbf{45.31} 
& \textbf{52.79} 
& \textbf{17.46} 
& \textbf{51.57} 
& \textbf{23.40} \\
\midrule
\multirow{2}{*}{P/PG}
& PathAgent & 39.84 & \textbf{47.21} & 10.06 & 51.63 & \textbf{17.80} \\
& \textbf{BEACON} 
& \textbf{45.31} 
& 45.69 
& \textbf{19.24} 
& \textbf{57.99} 
& 16.67 \\
\midrule
\multirow{2}{*}{C/PG}
& PathAgent & 43.75 & \textbf{47.72} & 14.09 & 51.71 & \textbf{17.08} \\
& \textbf{BEACON} 
& \textbf{44.53} 
& \textbf{47.72} 
& \textbf{17.98} 
& \textbf{59.46} 
& 16.67 \\
\bottomrule
\end{tabular}
\caption{Backbone compatibility of BEACON across pathology VLMs and MLLMs. P: PLIP, C: CONCH, R1: Patho-R1-7B, and PG: PathGen-LLaVA.}
\label{tab:backbone_ablation}
\end{table}

\section{Conclusion}
\label{sec:conclusion}

We introduce BEACON, a training-free agentic framework that reframes WSI reasoning as Bayesian evidence acquisition rather than relevance-driven retrieval. BEACON maintains probabilistic beliefs over diagnostic hypotheses and sequentially acquires evidence that maximizes expected information gain. Across five zero-shot WSI-VQA benchmarks and diverse foundation-model backbones, BEACON consistently improves reasoning accuracy and acquisition efficiency. Our analyses further reveal that semantic relevance is an unreliable proxy for diagnostic informativeness, highlighting the importance of modeling diagnostic uncertainty during evidence acquisition.

\bibliography{aaai2027}

\clearpage

\appendix

\begin{table*}[t]
\centering
\small
\renewcommand{\arraystretch}{0.9}
\begin{tabular}{lp{10cm}}
\toprule
\textbf{Notation} & \textbf{Definition} \\
\midrule

\multicolumn{2}{l}{\textbf{Problem Formulation}} \\
$\mathcal{X}$ & Set of $N$ WSI patches, $\mathcal{X} = \{x_n\}_{n=1}^{N}$ \\
$N$ & Number of image patches in a WSI \\
$c_x$ & Spatial coordinates of patch $x$ \\
$e_x$ & VLM embedding of patch $x$, $e_x \in \mathbb{R}^{d}$ \\
$d_x, \tilde{d}_x$ & Offline and question specific descriptions of patch $x$ \\
$q$ & Diagnostic question \\
$\mathcal{Y}$ & Set of $K$ answer candidates, $\mathcal{Y} = \{y_1, \ldots, y_K\}$ \\
$K$ & Number of answer candidates \\
$y^\ast$ & Ground truth answer \\
$f_{\mathrm{MLLM}}$ & MLLM used for question conditioned patch inspection \\
$\mathrm{mag}$ & Patch magnification level \\
\midrule

\multicolumn{2}{l}{\textbf{Sequential Evidence Acquisition}} \\
$t$ & Evidence acquisition round index \\
$\mathcal{S}_t, \mathcal{U}_t$ & Acquired and unobserved patch sets at round $t$ \\
\midrule

\multicolumn{2}{l}{\textbf{Stage 1: Coarse Candidate Retrieval}} \\
$r_t$ & Retrieval query at round $t$ \\
$\phi_T(\cdot)$ & Frozen VLM text encoder \\
$\phi_I(\cdot)$ & Frozen VLM image encoder \\
$s_t(x)$ & Cosine similarity between $r_t$ and patch $x$ \\
$\mathcal{C}_t$ & Top-$M$ candidate pool retrieved at round $t$ \\
\midrule

\multicolumn{2}{l}{\textbf{Stage 2: Belief Construction}} \\
$h_k$ & Hypothesis that answer candidate $y_k$ is correct \\
$f_{\mathrm{R}}$ & Frozen reranker used for hypothesis support estimation \\
$R_t(x,k), \bar{R}_t(x,k)$ & Raw and normalized support of patch $x$ for hypothesis $h_k$ \\
$\pi_t(k)$ & Belief that hypothesis $h_k$ is correct at round $t$ \\
\midrule

\multicolumn{2}{l}{\textbf{Stage 3: Bayesian Evidence Acquisition}} \\
$Y$ & Unknown ground truth diagnosis \\
$O_x$ & Binary observation obtained from patch $x$ \\
$H(\pi_t)$ & Entropy of the current belief distribution \\
$\mathrm{EIG}_t(x)$ & Expected information gain of inspecting patch $x$ \\
$\mathcal{B}_t$ & Top-$B$ patches selected by EIG at round $t$ \\
\midrule

\multicolumn{2}{l}{\textbf{Stage 4: Evidence Control}} \\
$\mathcal{E}_t$ & Evidence acquired at round $t$ \\
$\Psi$ & Evidence controller \\
$\mathrm{state}_t$ & Controller decision: \textsc{Sufficient}, \textsc{Continue}, or \textsc{Zoom} \\
$r_{t+1}$ & Refined retrieval query for the next round \\
\midrule

\multicolumn{2}{l}{\textbf{Hyperparameters}} \\
$M$ & Candidate pool size (default: 25) \\
$B$ & Number of acquired patches per round (default: 5) \\
$T$ & Maximum number of acquisition rounds (default: 5) \\
\bottomrule
\end{tabular}
\caption{Summary of the notations used in the main paper.}
\label{tab:notation}
\end{table*}

\begin{algorithm}[t]
    \caption{BEACON Inference Procedure}
    \label{alg:BEACON}
    \begin{algorithmic}[1]
        \small
        \REQUIRE WSI patches $\mathcal{X}$, question $q$, answer candidates $\mathcal{Y}$, maximum acquisition rounds $T$, candidate pool size $M$, patches per round $B$
        \ENSURE Predicted answer $\hat{y} \in \mathcal{Y}$

        \STATE $\mathcal{S}_1 \leftarrow \emptyset$, $\mathcal{U}_1 \leftarrow \mathcal{X}$, $r_1 \leftarrow q$
        \STATE Construct hypotheses $\{h_1, \ldots, h_K\}$ from $\mathcal{Y}$

        \FOR{$t = 1$ \TO $T$}

            \STATE \COMMENT{Stage 1: Coarse Candidate Retrieval}
            \STATE $\mathcal{C}_t \leftarrow \operatorname{TopM}_{x \in \mathcal{U}_t} \, s_t(x)$

            \STATE \COMMENT{Stage 2: Belief Construction}
            \STATE Compute $\bar{R}_t(x,k)$ for $x \in \mathcal{C}_t$, $k = 1, \ldots, K$
            \STATE Aggregate belief $\pi_t(k)$ for $k = 1, \ldots, K$

            \STATE \COMMENT{Stage 3: Bayesian Evidence Acquisition}
            \STATE Compute $\mathrm{EIG}_t(x)$ for $x \in \mathcal{C}_t$
            \STATE $\mathcal{B}_t \leftarrow \operatorname{TopB}_{x \in \mathcal{C}_t} \, \mathrm{EIG}_t(x)$
            \STATE $\mathcal{S}_{t+1} \leftarrow \mathcal{S}_t \cup \mathcal{B}_t$, \quad $\mathcal{U}_{t+1} \leftarrow \mathcal{U}_t \setminus \mathcal{B}_t$
            \STATE Obtain $\tilde{d}_x \leftarrow f_{\mathrm{MLLM}}(x,q,\mathcal{Y},c_x,\mathrm{mag})$ for $x \in \mathcal{B}_t$; $\mathcal{E}_t \leftarrow \{\tilde{d}_x \mid x \in \mathcal{B}_t\}$

            \STATE \COMMENT{Stage 4: Evidence Control}
            \STATE $(\mathrm{state}_t, r_{t+1}) \leftarrow \Psi(\mathcal{E}_t, q, \mathcal{Y})$

            \IF{$\mathrm{state}_t = \textsc{Sufficient}$}
                \STATE \textbf{break}
            \ELSIF{$\mathrm{state}_t = \textsc{Zoom}$}
                \STATE Perform zoom-in inspection and update $\mathcal{S}_{t+1}$; \textbf{break}
            \ENDIF

        \ENDFOR

        \STATE $\hat{y} \leftarrow$ answer synthesized from the final acquired patch set
        \STATE \textbf{return} $\hat{y}$

    \end{algorithmic}
\end{algorithm}

\begin{table}[t]
\centering
\small
\begin{tabular}{ll}
\toprule
\textbf{Component} & \textbf{Detail} \\
\midrule
\multicolumn{2}{l}{\textbf{Hardware}} \\
GPU & 2 $\times$ NVIDIA RTX A5000 (24\,GB each) \\
CPU & 64 CPU cores \\
RAM & 386\,GB \\
\midrule
\multicolumn{2}{l}{\textbf{Operating System}} \\
Host OS & Ubuntu 20.04 LTS \\
Containerization & Apptainer/Singularity \\
\midrule
\multicolumn{2}{l}{\textbf{Software}} \\
Python & 3.9 \\
PyTorch & 2.5.1 (CUDA 12.1) \\
Transformers & 4.51.0 \\
Accelerate & 1.10.0 \\
scikit-learn & 1.5.2 \\
SciPy & 1.13.1 \\
OpenSlide-Python & 1.4.2 \\
\midrule
\multicolumn{2}{l}{\textbf{External API}} \\
GPT-5 access & OpenRouter API \\
\bottomrule
\end{tabular}
\caption{Computing infrastructure used for all experiments.}
\label{tab:compute}
\end{table}

\section{Diagnostic Hypothesis Construction}
\label{sec:supp_hypothesis}

Stage 2 (Belief Construction) maintains one diagnostic hypothesis \(h_k\) for each answer candidate \(y_k \in \mathcal{Y}\). In the standard case, answer choices are converted directly into diagnostic hypotheses. Representing answer candidates as explicit hypotheses enables the subsequent belief update and expected information gain computation to operate over a well-defined, mutually exclusive hypothesis space rather than free-form text.
Most answer choices require no additional processing. This section describes the two special cases that require hypothesis construction rules: negation-style questions and composite answer choices.

\noindent\textbf{Negation-Style Questions.}
Some questions ask which finding is absent or least likely (e.g., ``Which of the following is \textbf{not} observed in the slide?''). We detect such questions using a predefined set of negation cues (e.g., ``absent,'' ``not observed,'' ``least evident,'' and ``which is not''). For these questions, the constructed hypothesis is inverted such that \(h_k\) states that the finding specified by \(y_k\) is absent or not observed. For example, given this question and the candidate answer ``necrosis,'' the constructed hypothesis states that necrosis is absent from the tissue rather than present. The reranker is subsequently instructed to predict ``yes'' when the visual evidence supports this absence. This design ensures that the support score \(R_t(x,k)\) is always interpretable as the degree to which patch \(x\) supports hypothesis \(h_k\), irrespective of question polarity.

\noindent\textbf{Composite Answer Choices.}
Some multiple-choice questions contain answer options such as ``All of the above'' or ``Both A and B,'' which do not correspond to a single visual finding. For these cases, we expand the answer choice into the corresponding combination of its constituent options before constructing \(h_k\). For example, if the composite option is ``All of the above'' and the preceding choices describe adenocarcinoma, squamous cell carcinoma, and small cell carcinoma, the expanded hypothesis asserts that visual evidence for all three findings is simultaneously present. Similarly, a ``None of the above'' option is expanded into the joint negation of every other listed finding, requiring the reranker to confirm the absence of all findings rather than the presence of any single alternative. By expanding composite options into concrete diagnostic findings, the reranker reasons over clinically meaningful hypotheses rather than abstract meta-labels. Table~\ref{tab:hypothesis_expansion} summarizes the supported expansion rules, with any answer choice that matches no predefined pattern used as-is. All hypotheses for a given case are constructed once, prior to Bayesian evidence acquisition, so subsequent rounds reuse the same $h_k$ without re-derivation.

\begin{table}[t]
\centering
\small
\begin{tabularx}{\columnwidth}{l X}
\toprule
\textbf{Answer Choice Pattern} & \textbf{Hypothesis Expansion} \\
\midrule
``All of the above'' & Conjunction of all remaining options \\
``None of the above'' & Negation of all remaining options \\
``Both $A$ and $B$'' & Conjunction of $A$ and $B$ \\
``$A$ or $B$'' / ``Either $A$ or $B$'' & Disjunction of $A$ and $B$ \\
``Neither $A$ nor $B$'' & Both $A$ and $B$ are false \\
``All except $A$'' & Conjunction of all remaining options excluding $A$ \\
\bottomrule
\end{tabularx}
\caption{Recognized composite answer-choice patterns and their corresponding hypothesis expansions. ``Remaining options'' refers to the other, non-composite answer choices for the same question.}
\label{tab:hypothesis_expansion}
\end{table}

\section{Hypothesis Support Scoring}
\label{sec:supp_scoring}
After constructing the diagnostic hypotheses, Stage 2 (Belief Construction) computes a support score for every candidate patch--hypothesis pair using a frozen reranker $f_{\mathrm{R}}$. The resulting scores are subsequently row-normalized and aggregated to form the round belief distribution $\pi_t$. This section describes the reranker prompt template and the procedure used to extract the yes/no support probability.

\subsection{Reranker Prompt Template}
Hypothesis support is computed using a frozen text-only reranker that takes a pathology evidence description and a candidate hypothesis as input and predicts whether the evidence supports the hypothesis. Table~\ref{tab:reranker_prompt} presents the exact prompt template. Given a candidate set $\mathcal{C}_t$ containing $M$ patches and $K$ diagnostic hypotheses, the reranker evaluates all $M \times K$ patch--hypothesis pairs in each acquisition round.

The candidate patches in $\mathcal{C}_t$ are drawn from the unobserved set $\mathcal{U}_t$ and have not yet been inspected by the question-aware perceptor. Consequently, hypothesis support is computed using the offline, question-agnostic pathology description $d_x$. Question-specific descriptions generated after a patch is acquired are not used during belief construction. Two special cases affect how the prompt is assembled:

\noindent\textbf{Negation-Style Questions.}
When the question contains negation cues (e.g., ``absent,'' ``not observed, and ``which is not''), the task instruction is replaced with the negation-aware variant. This ensures that a ``yes'' prediction still means the evidence supports the (negated) hypothesis $h_k$.

\noindent\textbf{Composite Answer Choices.}
Composite options such as ``All of the above'', ``Both $A$ and $B$'', or ``None of the above'' are expanded into concrete diagnostic statements \emph{before} the prompt is assembled (see Section~\ref{sec:supp_hypothesis}). Consequently, the category description $h_k$ already contains the expanded form, and the prompt template itself does not require an additional variant for composites.

As a result, a ``yes'' prediction is consistently interpreted as evidence supporting hypothesis $h_k$, regardless of question polarity or whether the original answer choice was composite.

\begin{table*}[t]
\centering \small
\begin{tabularx}{\textwidth}{l X}
\toprule
\textbf{Component} & \textbf{Content} \\
\midrule
(1) System instruction $I_{\text{sys}}$ &
\texttt{Judge whether the pathology evidence in <Evidence Description> supports the hypothesis in <Category Description>. Respond with only `yes' if the evidence supports the hypothesis, otherwise `no'.} \\
\midrule
\multicolumn{2}{l}{\textit{(2) Task instruction $I_{\text{task}}$}} \\
\midrule
Standard question &
\texttt{Assume the pathology option (hypothesis) is true. Evaluate how likely the observed evidence would be under this assumption. If the evidence is expected given the hypothesis, respond `yes'. If the evidence would be unlikely or contradictory, respond `no'.} \\
\addlinespace
Negation-style question &
\texttt{Assume the pathology option (hypothesis) is true. This is a negation-style pathology question, so a correct hypothesis means the named feature is absent, not observed, or less evident than the alternatives in the slide. Evaluate how likely the observed evidence would be under this assumption. If the evidence matches this negation-style hypothesis, respond `yes'. If the evidence instead suggests the feature is present or prominent, respond `no'.} \\
\midrule
\multicolumn{2}{l}{\textit{(3) Category description $h_k$}} \\
\midrule
Standard question &
\texttt{Pathology question: [QUESTION]}\newline
\texttt{Hypothesis: The correct answer is `[CHOICE]'.}\newline
\texttt{If this is true, the slide should show: [EXPANDED CHOICE TEXT]} \\
\addlinespace
Negation-style question &
\texttt{Pathology question: [QUESTION]}\newline
\texttt{Hypothesis: The correct answer is `[CHOICE]' because this feature is absent, not observed, or less evident than the alternatives in the slide.}\newline
\texttt{If this is true, the slide should indicate that `[EXPANDED CHOICE TEXT]' is missing, not supported by the evidence, or less evident than the other options.} \\
\midrule
\multicolumn{2}{l}{\textit{(4) Assembled reranker input (chat template)}} \\
\midrule
Full input &
\texttt{[System]}\newline
\texttt{[SYSTEM INSTRUCTION $I_{\text{sys}}$]}\newline
\texttt{[User]}\newline
\texttt{<Instruct>:[TASK INSTRUCTION $I_{\text{task}}$]}\newline
\texttt{<Evidence Description>:[EVIDENCE]}\newline
\texttt{<Category Description>:[$h_k$]} \\
\bottomrule
\end{tabularx}
\caption{Reranker prompt template used to compute the hypothesis support score $R_t(x,k)$.}
\label{tab:reranker_prompt}
\end{table*}

\subsection{Yes/No Probability Extraction}
The full input shown in Table~\ref{tab:reranker_prompt} follows the reranker's chat template: a system turn containing the fixed system instruction $I_{\text{sys}}$ and a user turn containing the assembled \texttt{<Instruct>/<Evidence Description>/<Category Description>} block. An empty reasoning span (\texttt{<think></think>}) is inserted immediately before generation to encourage the model to directly produce a single-token yes/no prediction.

Given the templated input, we read the logits at the final input position and retain only the vocabulary entries corresponding to the literal tokens \texttt{yes} and \texttt{no}. A two-way softmax over these logits yields the raw hypothesis support score $R_t(x,k)$. All $M \times K$ patch--hypothesis pairs are evaluated using batched forward passes (default batch size of 4), with each input truncated to a maximum sequence length of 4096 tokens.

\section{Evaluation Protocol}
\label{sec:supp_eval}

PathAgent evaluates generated responses using character-level string similarity between the model output and the ground-truth answer text. Because this heuristic measures surface-level textual overlap rather than the semantic answer selected by the model, it can produce incorrect evaluations when responses are expressed as option identifiers (e.g., ``A''), abbreviations, or free-form explanations.

Table~\ref{tab:pathagent_eval} presents a representative example from TCGA SlideBench. The model outputs ``A'' and justifies the prediction by describing ``dense fibrous stroma'' (choice A), whereas the correct answer is ``amyloid-like stroma'' (choice C). Under string similarity scoring, the character ``A'' partially matches the ground-truth text ``Amyloid-like stroma'', causing the prediction to be incorrectly marked as correct despite selecting the wrong answer. This occurs because the evaluation protocol compares raw text strings without first resolving the generated response to an explicit answer choice.

To address this limitation, we use an LLM-based evaluator that maps each generated response, whether an option identifier, abbreviation, or free-form explanation, to its intended answer choice before computing accuracy. By evaluating semantic answer selection rather than surface-level textual similarity, our protocol is format-agnostic and correctly identifies the example as incorrect.

\begin{table*}[t]
\centering
\small
\begin{tabularx}{\textwidth}{l X}
\toprule
\multicolumn{2}{l}{\textbf{Question and Model Prediction}} \\
\midrule
Dataset & TCGA SlideBench (Slide ID: TCGA-50-6591-01Z-00-DX1) \\
Question & Examine the stroma pattern in the provided slide and identify which pattern is often associated with poorly differentiated adenocarcinoma featuring clear cell change. \\
Answer Choices & A. Dense fibrous stroma, B. Myxoid stroma, C. Amyloid-like stroma, D. Osteoid-like stroma \\
Ground Truth & Amyloid-like stroma (C) \\
Generated Output & ``A'' \\
Generated Explanation & ``The dense fibrous stroma is associated with poorly differentiated adenocarcinoma featuring clear cell change...'' \\
Actual Correctness & \textbf{Incorrect} (the model selected option A rather than the ground truth option C) \\
\midrule
\multicolumn{2}{l}{\textbf{Evaluation Comparison}} \\
\midrule
PathAgent (string similarity) & Compares the generated output ``A'' directly with the ground truth text ``Amyloid-like stroma'' using character level string similarity; scored \textbf{Correct} \\
Ours (LLM based mapping) & Resolves the generated output and explanation as selecting Dense fibrous stroma (A), then compares it against the ground truth Amyloid-like stroma (C); scored \textbf{Incorrect} \\
\bottomrule
\end{tabularx}
\caption{Comparison of evaluation protocols on a representative example from TCGA SlideBench (Slide ID: TCGA-50-6591-01Z-00-DX1), where character level string similarity yields a false positive that our LLM based evaluator correctly resolves.}
\label{tab:pathagent_eval}
\end{table*}

\section{Component Ablation Settings}
\label{sec:supp_ablation}

We detail the component ablation settings used to isolate the contribution of each stage in BEACON. Each ablation removes the target component while keeping the remaining stages unchanged. Unless otherwise specified, all ablations use the default backbone configuration and hyperparameter settings ($M{=}25$, $B{=}5$, and $T{=}5$).

\begin{itemize}

    \item \textbf{w/o Stage 1 (Coarse Candidate Retrieval):} Stage 1 is removed by skipping the top-$M$ candidate retrieval step. Consequently, both Stage 2 (Belief Construction) and Stage 3 (Bayesian Evidence Acquisition) operate over the entire unobserved patch set $\mathcal{U}_t$ at each round rather than the retrieved candidate pool $\mathcal{C}_t$. All subsequent stages remain unchanged.

    \item \textbf{w/o Stage 3 (Bayesian Evidence Acquisition):} Stage 3 is removed while retaining Stage 2 (Belief Construction). Candidate patches are no longer ranked by expected information gain (EIG). Instead, each candidate in $\mathcal{C}_t$ is ranked by the Shannon entropy of its row normalized hypothesis support distribution $\bar{R}_t(x,\cdot)$, and the top-$B$ lowest entropy (i.e., most decisive) patches are selected for acquisition. This heuristic uses the same reranker outputs as BEACON but does not condition patch selection on the current belief distribution $\pi_t$ or simulate the expected posterior update.

    \item \textbf{w/o Stage 2+3 (Belief Construction and Bayesian Evidence Acquisition):} Both Stage 2 and Stage 3 are removed. Candidate patches are directly ranked by their VLM relevance scores $s_t(x)$, and the top-$B$ patches from $\mathcal{C}_t$ are acquired at each round. This setting is equivalent to the relevance-based retrieval baseline.

    \item \textbf{w/o Stage 4 (Evidence Control):} Stage 4 is removed by disabling the evidence controller $\Psi$. The retrieval query is fixed to the original question $q$ throughout inference, i.e., $r_{t+1}=q$ for all rounds. Query refinement, early termination, and zoom operations are disabled, and acquisition proceeds until the maximum round budget $T$ is exhausted. The final prediction is synthesized from all acquired evidence accumulated across the $T$ rounds.

\end{itemize}

\section{Hyperparameter Sensitivity: Coarse Candidate Pool Size ($M$)}
\label{sec:supp_hparam_m}

We evaluate the sensitivity of BEACON to the coarse candidate pool size $M$ used in Stage 1 (Coarse Candidate Retrieval; Section~3.3). At each acquisition round, Stage 1 retrieves the top-$M$ patches most relevant to the current retrieval query and passes them to Stages 2--3 for belief construction and Bayesian evidence acquisition. The main paper uses $M=25$ (Table~\ref{tab:notation}) across all experiments.

We sweep $M\in\{0,25,50,75,100\}$ while fixing all other hyperparameters ($B=5$, $T=5$), using the CONCH backbone on TCGA SlideBench. Setting $M=0$ disables Stage 1, causing Stages 2--3 to operate directly on the full unobserved patch set, equivalent to the ``w/o Stage 1'' ablation in Section~\ref{sec:supp_ablation}.

\begin{figure}[t]
\centering
\includegraphics[width=0.85\linewidth]{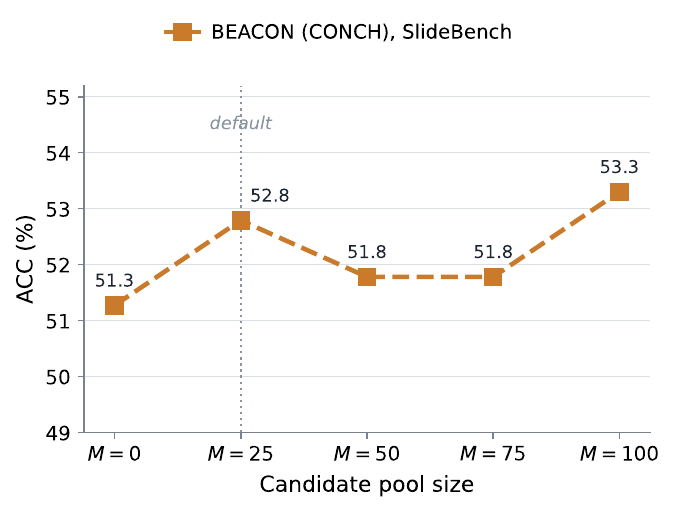}
\caption{Sensitivity of BEACON (CONCH) to the coarse candidate pool size $M$ on TCGA SlideBench.}
\label{fig:m_sweep}
\end{figure}

Figure~\ref{fig:m_sweep} shows that BEACON is robust to the choice of $M$, with accuracy remaining within a narrow range (51.3--53.3\%) across the sweep. The lowest accuracy is obtained at $M=0$ (51.3\%), suggesting that Stage 1 provides a modest but consistent performance benefit.

\section{Analysis Settings}
\label{sec:supp_why}

This section details the experimental settings used to compare Bayesian (EIG based) and relevance based patch acquisition under identical conditions.

\noindent\textbf{Experimental Setup.} All analyses use the default backbone configuration (PLIP + Qwen3-Reranker-4B + Patho-R1-7B + Qwen3-4B) and are conducted on 325 cases from TCGA ExpertVQA (128) and TCGA SlideBench (197), yielding 329 total evidence acquisition rounds. The additional rounds arise from cases that require more than one acquisition round before reaching \textsc{Sufficient} or \textsc{Zoom}. For every acquisition round, we compare BEACON's EIG selected patch batch with a counterfactual relevance selected batch constructed from the identical prior belief distribution $\pi_t$, candidate pool $\mathcal{C}_t$ ($M{=}25$), and reranker support matrix $R_t$. The only difference between the two batches is the patch acquisition strategy used for selection (EIG versus relevance score $s_t(x)$).

\noindent\textbf{Rank Correlation Analysis.} For each acquisition round, candidate patches in $\mathcal{C}_t$ are independently ranked by relevance score $s_t(x)$ and expected information gain $\mathrm{EIG}_t(x)$. We compute the Spearman rank correlation coefficient between the two rankings and report its distribution over all 329 acquisition rounds.

\noindent\textbf{Expected Uncertainty Reduction Analysis.} For each acquisition round, we compute the normalized expected reduction in diagnostic uncertainty, $\mathrm{EIG}(\mathcal{B})/H(\pi_t)$, for the EIG selected and relevance selected top-$B$ patch batches. Both batches are constructed from the same candidate pool and prior belief distribution. The distribution of normalized uncertainty reduction is reported across all 329 acquisition rounds.

\noindent\textbf{Uncertainty Stratified Batch Diversity Analysis.} Batch diversity is defined as one minus the mean pairwise cosine similarity between the normalized hypothesis support vectors $\bar{R}_t(x,\cdot)$ of the selected patches. Higher diversity indicates that selected patches provide complementary diagnostic evidence by supporting different hypotheses. Acquisition rounds are stratified into quartiles based on prior diagnostic uncertainty $H(\pi_t)$, with quartile boundaries computed from the empirical distribution. We report the mean batch diversity together with 95\% confidence intervals for both EIG selected and relevance selected patch batches within each uncertainty quartile.

\section{Token Usage and API Cost}
\label{sec:supp_tokens}

Table~\ref{tab:openrouter_tokens} reports the full input/output token counts and estimated API cost for GIANT, PathAgent, and BEACON under the GPT-5 backbone across all five MultiPathQA benchmarks, computed at OpenRouter list pricing (\$1.25 / \$10 per 1M input/output tokens). All costs are computed from raw token counts at list pricing, without prompt-caching discounts, and are therefore an upper bound on actual billed cost.

\begin{table}[t]
\centering
\small
\setlength{\tabcolsep}{1pt}
\begin{tabular}{llrrrr}
\toprule
\textbf{Method} & \textbf{Dataset} & \textbf{\#Cases} & \textbf{In Tok.} & \textbf{Out Tok.} & \textbf{Cost(\$)} \\
\midrule
\multirow{6}{*}{GIANT}
& ExpertVQA  & 128 & 14{,}110{,}057 &   670{,}338 & 24.34 \\
& SlideBench & 197 & 22{,}537{,}014 & 1{,}108{,}821 & 39.26 \\
& TCGA       & 221 & 25{,}532{,}372 & 1{,}170{,}494 & 43.62 \\
& GTEx       & 190 & 20{,}909{,}732 &   923{,}118 & 35.37 \\
& PANDA      & 196 & 22{,}277{,}687 & 1{,}281{,}623 & 40.66 \\
& \textbf{Total} & 932 & 105{,}366{,}862 & \textbf{5{,}154{,}394} & 183.25 \\
\midrule
\multirow{6}{*}{PathAgent}
& ExpertVQA  & 128 & 3{,}040{,}175 & 2{,}669{,}971 & 30.50 \\
& SlideBench & 197 & 5{,}055{,}521 & 4{,}872{,}745 & 55.05 \\
& TCGA       & 221 & 5{,}338{,}371 & 4{,}451{,}973 & 51.19 \\
& GTEx       & 190 & 1{,}110{,}094 & 2{,}347{,}688 & 24.86 \\
& PANDA      & 196 & 1{,}057{,}701 & 2{,}797{,}950 & 29.30 \\
& \textbf{Total} & 932 & 15{,}601{,}862 & 17{,}140{,}327 & 190.91 \\
\midrule
\multirow{6}{*}{BEACON}
& ExpertVQA  & 128 & 569{,}933 & 1{,}368{,}493 & 14.40 \\
& SlideBench & 197 & 1{,}571{,}715 & 2{,}585{,}315 & 27.82 \\
& TCGA       & 221 & 1{,}334{,}412 & 2{,}720{,}955 & 28.88 \\
& GTEx       & 190 & 727{,}173 & 1{,}257{,}925 & 13.49 \\
& PANDA      & 196 & 798{,}685 & 1{,}612{,}052 & 17.12 \\
& \textbf{Total} & 932 & \textbf{5{,}001{,}918} & 9{,}544{,}740 & \textbf{101.70} \\
\bottomrule
\end{tabular}
\caption{Token usage and estimated API cost (OpenRouter \texttt{gpt-5} list pricing, \$1.25/\$10 per 1M in/out tokens) on MultiPathQA, with GIANT run for 20 iterations. All costs are computed from raw token counts at list pricing, without prompt-caching discounts, so reported values are an upper bound on actual billed cost.}
\label{tab:openrouter_tokens}
\end{table}


\end{document}